# Nonlinear Model Predictive Control with Single-Shooting Method for Autonomous Personal Mobility Vehicle

**Rakha Rahmadani Pratama** [a,b], **Catur Hilman A.H.B. Baskoro** [a], **Joga Dharma Setiawan** [b], **Dyah Kusuma Dewi** [a], **P Paryanto** [b], **Mochammad Ariyanto** [c], **Roni Permana Saputra** [a, *]

[a] *Research Center for Smart Mechatronics, National Research and Innovation Agency (BRIN), Jalan Sangkuriang, Bandung, 40135, Indonesia*

[b] *Department of Mechanical Engineering, Diponegoro University, Jl. Prof. Soedarto No.13, Tembalang, Kota Semarang, 50275, Indonesia*

[c] *Department of Mechanical Engineering, Osaka University, 1-1 Yamadaoka, Suita, Osaka, 565-0871, Japan*

*Corresponding Author: roni.permana.saputra@brin.go.id*

**Abstract**

This paper introduces a proposed control method for autonomous personal mobility vehicles, specifically the Single-passenger Electric Autonomous Transporter (SEATER), using Nonlinear Model Predictive Control (NMPC). The proposed method leverages a single-shooting approach to solve the optimal control problem (OCP) via non-linear programming (NLP). The proposed NMPC is implemented to a non-holonomic vehicle with a differential drive system, using odometry data as localization feedback to guide the vehicle towards its target pose while achieving objectives and adhering to constraints, such as obstacle avoidance. To evaluate the performance of the proposed method, a number of simulations have been conducted in both obstacle-free and static obstacle environments. The SEATER model and testing environment have been developed in the Gazebo Simulation and the NMPC are implemented within the Robot Operating System (ROS) framework. The simulation results demonstrate that the NMPC-based approach successfully controls the vehicle to reach the desired target location while satisfying the imposed constraints. Furthermore, this study highlights the robustness and real-time effectiveness of NMPC with a single-shooting approach for autonomous vehicle control in the evaluated scenarios.

# I. Introduction

In the era of rapid technological development, autonomous vehicles are one of the most transformative technologies set to revolutionize the transportation sector [1]. The development of autonomous vehicles is continuously being advanced by the industry and scientific community to enable new mobility concepts, improve road safety, and achieve higher automation [2]. Reliable and effective control approaches are essential to fully realize the potential of autonomous vehicles.

Model Predictive Control (MPC) is one of the prominent control approaches that has gained significant attention. Over the last decades, MPC has shown significant success in high-performance control of complex systems such as industrial processes, intelligent building operations, automation, power electronics, and autonomous robotic systems [3]-[7]. Additionally, MPC is capable of handling challenging conditions such as adverse weather [8] and addressing non-linearities while supporting Multi-Input–Multi-Output (MIMO) systems [9]. By enabling predictive control, MPC facilitates timely and responsive decision making, enhancing the autonomous vehicle's ability to navigate dynamic environment.

This study leverages the MPC technique to develop a control approach for the Single-passenger Electric Autonomous Transporter (SEATER) platform. Using the non-linear Model Predictive Control (NMPC) method, the SEATER vehicle is expected to achieve improved control performance, respond adaptively to different situations, and ensure accurate positioning control. More specifically, the contributions of this study are as follows:

1. The design and formulation of an NMPC approach for addressing the SEATER platform's navigation problem, including waypoint path following.
2. The development and modelling of the SEATER platform in the Gazebo physics engine to conduct simulation experiments and validate the proposed NMPC approach.
3. The integration of the NMPC controller within the Robot Operating System (ROS) framework.

The remainder of the paper is organized as follows: Section II provides an overview of the related work relevant to this study. Section III presents the NMPC's problem formulation and implementation of the NMPC on the SEATER platform. Section IV discusses the simulation experiments and analyses the results. Finally, Section V summarizes the findings.

# II. Related Work

MPC is one of the widely used control approaches in dynamic system control, primarily due to its ability to handle multivariable systems and consider physical and environmental constraints, leveraging optimization techniques [10]. The MPC methods have developed rapidly along with the advanced in various numerical approaches used to solve the optimization problems [11].

Optimal control problems are typically addressed using two main approaches: direct and indirect techniques [12]. The direct method transforms the optimal control problem into nonlinear programming (NLP) through suitable discretization techniques [13]. Over the past few decades, numerous direct approaches have been established, including the shooting method. Single shooting in MPC has been widely applied in various fields, including the control of non-holonomic robots. One of the main advantages of this method is its simplicity in implementation and the ability to integrate complex system models with relative ease [14]. In the context of autonomous vehicles, one of the challenges lies in maintaining stability of the vehicle motion while producing efficient trajectories. MPC with single shooting could effectively addresses this challenge by providing an optimal solution that accounts for the complex dynamics of the robot.

The study presented in [15] applied single shooting techniques for controlling mobile robots in dynamic environments, where MPC enables the robots to rapidly adjust its trajectory in response to changing conditions or the emergence of unexpected obstacles. The single-shooting method offers advantages in terms of computation speed, as it only requires one optimization problem for each prediction period. Additionally, experimental testing on real-world vehicles, as presented in [2], highlights the capability of the NMPC method to effectively address complex and challenging tasks in real-world settings.

Our prior work in [16] introduced a non-linear model predictive control (NMPC)-based visual servoing method for autonomous docking of the SEATER platform, a single-seat personal mobility vehicle. This approach leverages real-time visual sensor feedback to guide the vehicle's motion. The simulation results demonstrate the successful generation of real-time docking trajectories across a range of scenarios. Expanding on this foundation, the current paper extends the problem formulation by incorporating waypoint trajectory following into the NMPC framework. Furthermore, the vehicle model and control system are implemented in the Gazebo physics engine to facilitate more realistic simulations that account for the dynamic properties of the vehicle model.

## III. Problem Formulation

### A. SEATER Kinematic Model and Model Discretization

SEATER is a single-passenger autonomous vehicle equipped with a differential drive system, comprising two independently controlled driving wheels. This configuration enables precise control of both speed and direction. By varying the speed of each independent wheel, various trajectories can be achieved [15]. Due to the inherent motion constraints of the differential drive system, SEATER is classified as a non-holonomic system, which limits its movement in certain directions and complicates its control compared to holonomic systems. Non-holonomic systems refer to systems that cannot move in all directions due to constraints such as wheel configuration, making their control more complex and requiring particular control strategies such as model predictive control (MPC) for accurate manoeuvring [11].

To formulate the model and control of SEATER, the system's state is represented in a continuous-time state space defined by the function f and the control input $u(t)$. This continuous-time model can be formulated in the form of a state space model, as shown in equation (1).

$$\begin{bmatrix} \dot{x} \\ \dot{y} \\ \dot{\theta} \end{bmatrix} = \begin{bmatrix} \cos(\theta) & 0 \\ \sin(\theta) & 0 \\ 0 & 1 \end{bmatrix} \begin{bmatrix} v \\ \omega \end{bmatrix} \quad (1)$$

This continuous model allows for a detailed representation of the vehicle's behavior, but to implement control algorithms effectively, a discrete-time model is often more practical. The Euler method is used to discretize the SEATER model, converting the continuous model into a form suitable for digital computation and control [15]. The reformulated discrete system equations are provided in equation (2).

$$x(k+1) = f(x(k), u(k)) \quad (2)$$

Eulerian discretization goes further by numerically solving ordinary differential equations (ODEs), as shown in equation (3).

$$\begin{bmatrix} x(k+1) \\ y(k+1) \\ \theta(k+1) \end{bmatrix} = \begin{bmatrix} x(k) \\ y(k) \\ \theta(k) \end{bmatrix} + \Delta T \begin{bmatrix} v(k)\cos(\theta(k)) \\ v(k)\sin(\theta(k)) \\ \omega(k) \end{bmatrix} \quad (3)$$

This method is applied to a system with three state variables, $x(k)$, $y(k)$, and $\theta(k)$. These variables represent an object's position $x, y,$ and orientation $\theta$ in two-dimensional space. Using this approach, the next state $(k+1)$ of the robot or vehicle can be predicted based on the current state at the time $k$. The calculation adds the product of the sampling time $\Delta T$, linear velocity $v$, and the trigonometric components $\cos(\theta(k))$ and $\sin(\theta(k))$ to the current position, enabling the model to estimate the vehicle's updated state.

**B. NMPC Cost Function**

Model Predictive Control (MPC) is an advanced control strategy that optimizes a system's future behavior by solving an optimization problem at each control step. MPC involves using a system dynamics model to predict a sequence of control inputs over a time interval known as the predictive horizon [17]. The control inputs then optimize a predetermined objective function with given constraints. Nonlinear model predictive control (NMPC) effectively manages complex nonlinear systems with multiple objectives and constraints [10]. For this reason, NMPC is chosen as the control strategy for SEATER.

To determine the optimal control sequence that produces a state $x(k)$ close to the reference value (set point) $x^r$ for $k = 0, \ldots, N-1$, the cost function or minimization of the cost function is used [15]. This cost function calculates the error of the current state $x(k)$ with the reference $x^r$ and the value of $u$, each of which has its square norm calculated and added together. as written in equation (4). In addition, it can provide corrections to the deviation of the control value $u(k)$ from the reference control. A weighting matrix with $Q$ (state cost weighting matrix) and $R$ (control effort weighting matrix) is also added.

$$l(x, u) = \|x_u - x^r\|_Q^2 + \|u - u^r\|_R^2 \tag{4}$$

The weighting matrix in the cost function can be fine-tuned to adjust the control performance [18]. In this study, the weighting matrices $Q$ and $R$ were defined as fixed variables, with $Q$ set such that $Q[0,0] = 1$, $Q[1,1] = 5$, $Q[2,2] = 0.1$, and R set such that $R[0,0] = 0.5$, $R[1,1] = 0.05$.

The MPC formulation integrates motion planning and trajectory tracking, generating the control input based on the Optimal Control Problem (OCP) problem [19]. The OCP aims to determine the control sequence that minimizes the cost function [20]. In this work, the OCP is formulated and solved as a Non-Linear Programming (NLP) problem. The standard NLP framework for numerical parametric optimization is formulated in equation (5).

$$\begin{aligned}
\text{minimize} \; : \; & J_N(x, u) = \sum_{k=0}^{N-1} l(x(k), u(k)) \\
\text{subject to} \; : \; & x_u(k+1) = f(x_u(k), u(k)), \\
& x_u(0) = x_0, \\
& u(k) \in U, \quad \forall k \in [0, N-1], \\
& x_u(k) \in X, \quad \forall k \in [0, N]
\end{aligned} \tag{5}$$

According to equation (5), the cost function to be minimized, $J$, represents the cumulative sum of the stage costs $l(x(k), u(k))$ at each time step $k$. This function quantifies the penalty or cost associated with the state $x(k)$ and the control input $u(k)$. The system must adhere the dynamics defined by $x_u(k+1)$ and start from an initial state $x_u(0)$. Furthermore, the control $u(k)$ and the state $x_u(k)$ must remain within their feasible sets $U$ and $X$, respectively, to ensure that the resulting solution is valid and satisfies all imposed constraints. While large prediction horizons can lead to better results, they come at the cost of increased computational demand. Therefore, a computational balance is required to select a prediction horizon that provides optimal performance while maintaining computational efficiency [21].

## C. NMPC Shooting Method

The shooting method in NMPC is a numerical approach used to solve optimal control problems by discretizing nonlinear dynamical systems and iteratively solving differential equations using ODE solvers and optimizing control inputs over a prediction horizon. This method allows modelling complex systems while applying constraints and increasing computational efficiency compared to other methods. These properties make it a practical approach for solving optimal control problems in nonlinear dynamical systems [15].

The single shooting method in NMPC uses a single trajectory to solve the MPC objective function and obtain the optimal solution. This method is straightforward, relying on a single sequence of control actions. Thus, this study uses single shooting method to solve the optimization problem for the NMPC. Single-shooting optimization involve a lower dimensional variable than its alternatives, therefore helping to speed up the solution process [14].

Let $w$ denote the decision variable of the optimization process, representing the control vector comprising a sequence of control values from $u_0$ to $u_{N-1}$ over the finite horizon N. These control values are the variables optimized in MPC.

$$w = [u_0, \dots, u_{N-1}] \tag{6}$$

The robot state trajectory $X_u(.)$ along the horizon $N$ can be expressed as a recursive function of the control trajectory using the robot's dynamic function.

$$\begin{aligned} X_u(.) &= F(W, X_0, t_k) \\ F(w, x_0, t_0) &= x_0 \end{aligned} \tag{7}$$

where $X_u(.)$ represents the state trajectory, dependent on the control vector $W$, initial state $X_0$, and time $t_k$, $F(W, X_0, t_k)$ describes the system's dynamics, showing how the system's state changes based on the applied control.

$$\min_w \Phi\left(F(w, x_0, t_k), w\right)$$
$$\text{subject to} : (F(w, x_i, t_k), w) \leq 0 \tag{7}$$

The expressions in equation (6) represent the optimization problem where the goal is to minimize a cost function while satisfying system constraints. These constraints could include physical, safety, and operational requirements that the system must adhere to. In this study to incorporate obstacle avoidance constraint, the obstacle area is limited by a circular boundary as formulated in equation (9).

$$\sqrt{(x - x_{ob})^2 + (y - y_{ob})^2} - \text{r} + r_{ob} \geq 0 \tag{8}$$

This constraint ensures that the distance between the system's position $(x, y)$ and the obstacle center $(x_{ob}, y_{ob})$ is always greater than or equal to the obstacle radius $(r_{ob})$. This condition defines a safe boundary around the obstacle, ensuring that the system remains outside this boundary and follows a secure path.

Algorithm 1 summarizes the proposed NMPC approach presented in this study. The process starts with defining the prediction horizon, initial state, and target state. At each sampling instance, the current state is estimated, and an OCP is solved to find the optimal control sequence that minimizes the cost function while satisfying system constraints. The first control input from the optimized sequence is then applied to the system, and this loop continues at each time step to guide the system towards the desired state effectively.

---

Algorithm 1: Nonlinear Model Predictive Control

---

**MPC Init:**

    Prediction Horizon:= N

    Define the initial vehicle state:

    $x_u(0) := x_0$

    Define the target state: $x^r$

    Define the initial control: $u_0$

    Apply $u_0$ to the system.

for every sampling instant k = 1, 2, ... do

    Estimate the states $x(k)$

    **Solve OCP:**

    Find the optimal control horizon

$$w = [u_0, \ldots, u_{N-1}]$$

    which satisfies

$$J_N(\hat{x}, u^*) = V_N$$

    s.t.

$$(F(w, x_i, t_k), w) \leq 0$$

    Apply $u_0$ to the system.

## IV. Results and Discussion

To evaluate the performance of the proposed NMPC approach for autonomous personal mobility vehicles, particularly the SEATER platform, a series of experimental scenarios were designed and implemented using Python and Gazebo simulation environments. The computational hardware employed for the simulations consisted of an Intel Core i7 processor with eight cores operating at 2.30 GHz. The experimental setup focused on scenarios involving static obstacles and flat terrain to ensure controlled and reproducible testing conditions. The NMPC algorithm was developed in Python, leveraging the CasADi framework for solving Nonlinear Programming (NLP) problems through algorithmic differentiation [22]

Table 1 MPC Data Collection with Obstacles-Free

| Parameter | | Performance | | | |
|---|---|---|---|---|---|
| $\Delta T$ | N | Total Time Iteration (s) | Max Time Iteration (s) | Euclidean Position Error (m) | Rotation Error (rad) |
| 0.01 | 5 | 10.002 | 0.007 | 1.086 | 0.021 |
| | 10 | 9.836 | 0.009 | 0.618 | 0.039 |
| | 15 | 9.031 | 0.017 | 0.222 | 0.004 |
| | 20 | 8.699 | 0.016 | 0.150 | 0.004 |
| | 25 | 8.295 | 0.016 | 0.129 | 0.010 |
| 0.05 | 5 | 10.165 | 0.014 | 0.167 | 0.018 |
| | 10 | 6.961 | 0.015 | 0.079 | 0.003 |
| | 15 | 6.959 | 0.020 | 0.059 | 0.009 |
| | 20 | 6.963 | 0.015 | 0.051 | 0.008 |
| | 25 | 6.996 | 0.019 | 0.043 | 0.009 |
| 0.1 | 5 | 6.818 | 0.014 | 0.084 | 0.025 |
| | 10 | 6.206 | 0.014 | 0.041 | 0.016 |
| | 15 | 1.523 | 0.017 | 0.039 | 0.014 |
| | 20 | 0.897 | 0.018 | 0.036 | 0.026 |
| | 25 | 0.802 | 0.026 | 0.034 | 0.028 |
| 0.5 | 5 | 0.171 | 0.015 | 0.026 | 0.014 |
| | 10 | 0.222 | 0.026 | 0.021 | 0.028 |
| | 15 | 0.361 | 0.029 | 0.021 | 0.004 |
| | 20 | 0.390 | 0.032 | 0.025 | 0.019 |
| | 25 | 0.452 | 0.041 | 0.026 | 0.021 |

The initial setup involved programming the SEATER in python simulations to move from the starting point (0, 0) to three target points: (1.5, 1.5), (1.5, 0), and (1.5, -1.5). The total iteration time was set at 10 seconds, with a maximum iteration time constrained to be less than the sampling time. The performance thresholds included a target Euclidean distance of 0.4 meters and a rotation threshold of 0.4 radians. Control noise of 10% and a localization error of 0.02 m were introduced to add realism to the tests. Simulations were conducted in two primary conditions: obstacle-free and static obstacle scenarios.

Table I presents the performance of NMPC under obstacle-free conditions. The results show that a shorter sampling time correlates with an increase in total iteration time, maximum duration, and error in distance and rotation that may lead to failure. The sampling time parameter of 0.5 seconds has satisfactorily fulfilled all standards. Items marked in grey do not satisfy the criteria, whereas those underlined parameters represent the failing factors.

Table 2 MPC Data Collection with Static Obstacles

| Parameter | | Performance | | | |
|---|---|---|---|---|---|
| $\Delta T$ | N | Total Time Iteration (s) | Max Time Iteration (s) | Euclidean Position Error (m) | Rotation Error (rad) |
| 0.10 | 15 | 9.897 | 0.030 | 0.914 | 0.284 |
| | 20 | 4.912 | 0.033 | 0.787 | 0.007 |
| | 25 | 7.317 | 0.042 | 0.784 | 0.016 |
| 0.50 | 5 | 3.566 | 0.023 | 0.796 | 0.027 |
| | 10 | 3.426 | 0.032 | 0.781 | 0.014 |
| | 15 | 3.480 | 0.039 | 0.765 | 0.032 |
| | 20 | 0.830 | 0.055 | 0.017 | 0.015 |
| | 25 | 0.820 | 0.075 | 0.018 | 0.021 |

The obstacle condition was then tested using parameters that met these thresholds in the previous experiments. The SEATER was modeled with a perimeter diameter of 0.3 m, and obstacles were positioned at (0.8, 0.3), (0.8, -0.3), and (1, 0), each with a 0.2 m diameter and a safety tolerance of 0.05 m to prevent intersection. Table II shows that the optimal performance was achieved with a sampling time of 0.5 seconds and prediction horizons of 20 and 25, with N = 20 identified as the most effective setting.

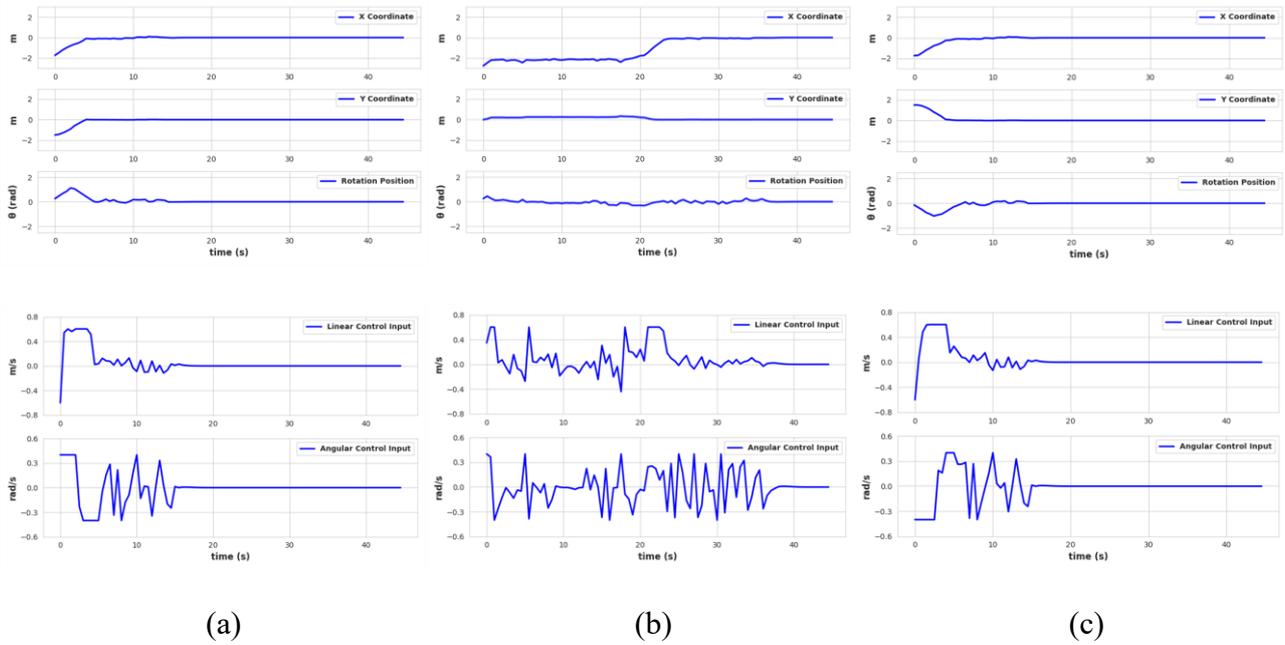

(a)                  (b)                  (c)

Figure 1. MPC Performance with static obstacle: (a) turning left; (b) straight; (c) turning right

Figure 1 illustrates the control performance and position history of the Python obstacle simulation under optimal parameters. It shows that it satisfies the specified speed constraint and reaches the destination with a final speed of zero, indicating its stability.

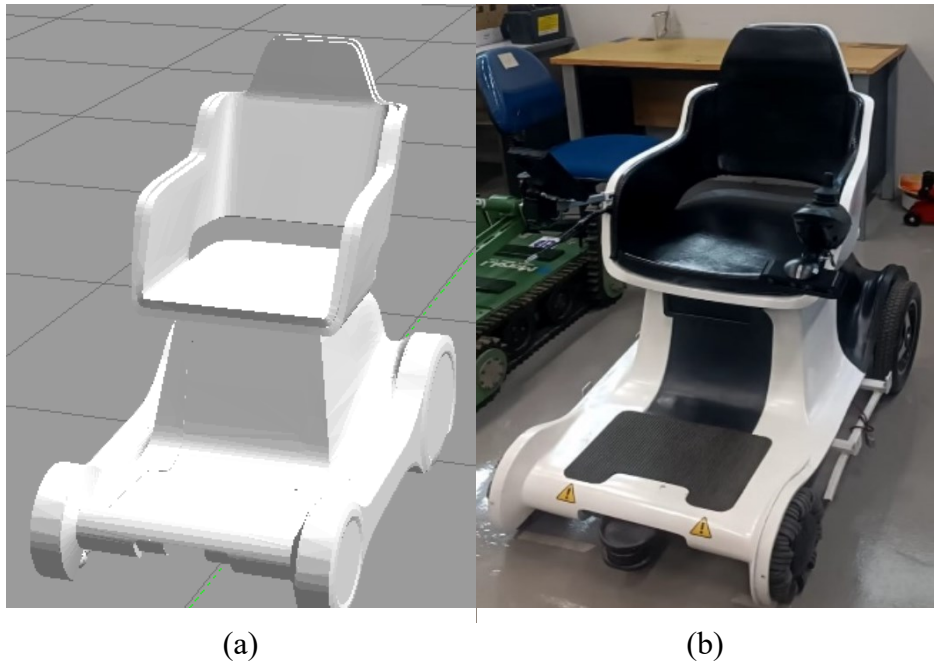

(a)                  (b)

Figure 2. SEATER model: (a) in Gazebo; (b) actual platform

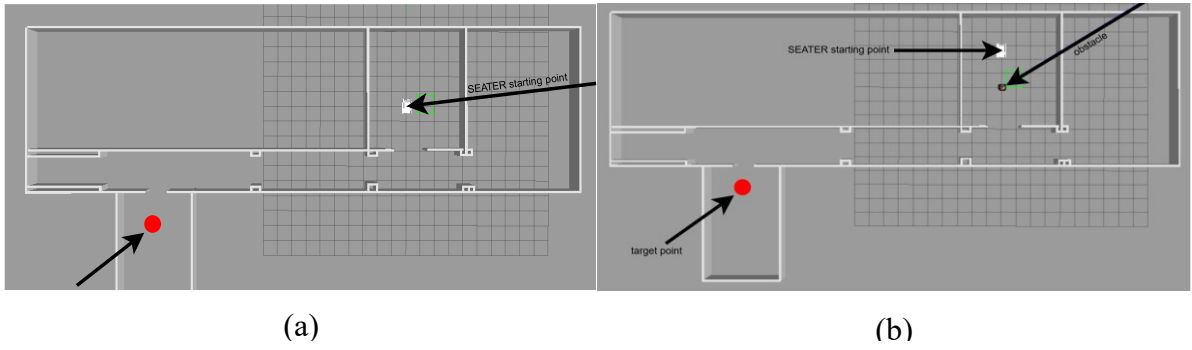

(a)　　　　　　　　　　　　　　　(b)

Figure 4 Simulation map: (a) obstacle-free; (b) static-obstacle

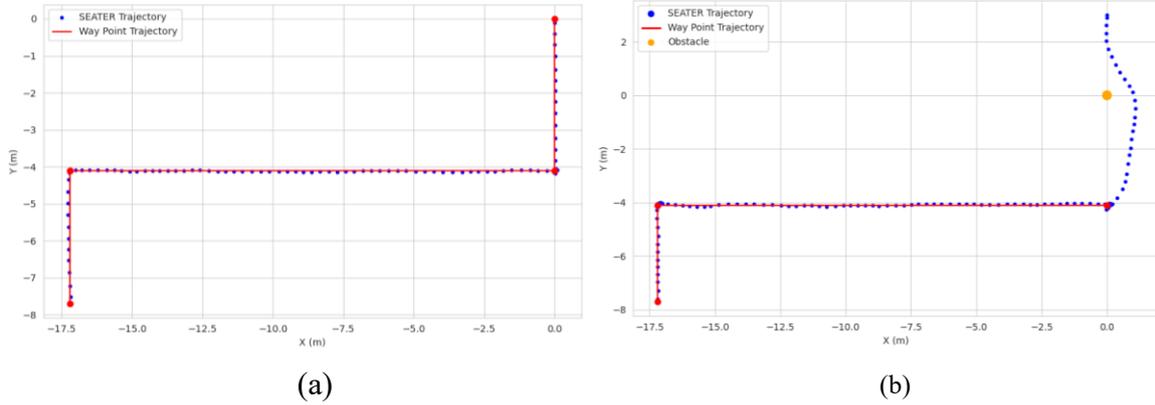

(a)　　　　　　　　　　　　　　　(b)

Figure 3. Comparison The SEATER Trajectory and The Waypoint Trajectory: (A) Obstacle-Free; (B) Static-Obstacle

The optimal parameters were then applied in Gazebo simulations within the ROS framework. Figures 2(a) and 2(b) illustrate the SEATER model in Gazebo and its real-world counterpart. The testing was performed on two maps: one without obstacles and one with static obstacles, as shown in Figures 3(a) and 3(b). SEATER's position data was recorded using odometry and compared to the planned trajectory. A control noise of 15% and a localization error of 0.04 m were applied to ensure realistic results.

Table 3 Error Trajectory and Error Final Seater Position For Obstacle-Free

| Number | Euclidean Position Error (m) | Rotation Error (rad) | Maximum Trajectory Error (m) | Average Trajectory error (m) |
|---|---|---|---|---|
| 1 | 0.049 | 0.105 | 0.223 | 0.057 |
| 2 | 0.077 | 0.075 | 0.109 | 0.046 |
| 3 | 0.099 | 0.136 | 0.105 | 0.028 |
| Average | 0.075 | 0.105 | 0.145 | 0.044 |

Table 4 Error Trajectory and Error Final Seater Position For Static-Obstacle

| Number | Euclidean Position Error (m) | Rotation Error (rad) | Maximum Trajectory Error (m) | Average Trajectory Error (m) | Minimum Euclidean Obstacle (m) |
|---|---|---|---|---|---|
| 1 | 0.054 | 0.142 | 0.099 | 0.033 | 0.758 |
| 2 | 0.108 | 0.104 | 0.114 | 0.024 | 0.825 |
| 3 | 0.096 | 0.064 | 0.149 | 0.029 | 0.896 |
| Average | 0.086 | 0.103 | 0.121 | 0.028 | 0.758 |

In the obstacle-free simulation, the SEATER moved from (0, 0, 1.57) through waypoints at (0, 4.1, 1.57), (0, 4.1, 0), (17.2, 4.1, 0), (17.2, 4.1, 1.57), and (17.2, 7.7, 1.57). The coordinates used are (x, y, and θ), where θ is the angular SEATER rotated in radians. Figure 3(a) compares the SEATER's trajectory to the planned path, with Table III reporting an average Euclidean position error of 0.075 meters and an average trajectory error of 0.044 meters across three trials. For the obstacle condition, the SEATER started at (0, 3, -1.57) with an obstacle positioned at (0, 0). Figure 3(b) shows the comparison between the SEATER's trajectory and the planned path, and Table IV summarizes the results, indicating an average minimum Euclidean distance of 0.758 meters from the obstacle.

The results confirm that the prediction horizon (N) and sampling time (ΔT) parameters identified in Python simulations were effectively implemented in the Gazebo environment, demonstrating the robustness of the NMPC approach.

## V. CONCLUSION

This paper presents a proposed control method for an autonomous personal mobility vehicle platform, SEATER, using Nonlinear Model Predictive Control (NMPC) with a single-shooting approach. The NMPC method is implemented and evaluated in Gazebo simulation environment. The NMPC single-shooting simulation experiment results show that small sampling time and prediction horizon values can hinder the control's ability to reach the desired target. Therefore, careful selection of the sampling time and prediction horizon is necessary to balance optimal performance and computational efficiency.

This study identifies that optimal performance is achieved with a sampling time of ΔT=0.5 seconds and a prediction horizon of N=20. In the Gazebo simulations, these parameters enabled the SEATER platform to effectively reach target positions and follow designated trajectories. Performance was assessed using metrics, including Euclidean position error, rotation error, average and maximum trajectory errors, and obstacle avoidance capability. Additionally, the proposed control method demonstrated adaptability under noise and localization disturbances, showcasing robustness in varied uncertainties.

Future research should explore the inclusion of additional control parameters and extend simulations to more complex environments to further enhance system performance and adaptability. Additionally, implementing and testing the proposed control method in real-world scenarios would be essential to validate its robustness and effectiveness.

## Acknowledgement

The experiments conducted in this study result from a collaboration between BRIN and Diponegoro University, which was fully funded by the Research Organization for Electronics and Informatics, National Research and Innovation Agency (BRIN).